%% file: kepafeng.tex
\documentclass[runningheads]{eng}

\usepackage{epsfig} % file "epsfig.sty" contains macros to include
\usepackage{float}
\usepackage{subcaption}
\usepackage{color}
\usepackage{colortbl}
\usepackage{relsize}
\usepackage{graphicx}
\usepackage[export]{adjustbox}
\usepackage{multirow}

\usepackage{todonotes}

\begin{document}
\frontmatter
\pagestyle{headings}
%\tableofcontents
\mainmatter

%%%%%%%%%%%%%%%%%%%%%%%%%%%%%%%%%%%%%%%%%%%%%%%%%%%%%%%%%%%

%\input paper1.tex    %   sample paper 1
\input waste_detection.tex     %%% your contribution
%\input paper2.tex    %   sample paper 2

%%%%%%%%%%%%%%%%%%%%%%%%%%%%%%%%%%%%%%%%%%%%%%%%%%%%%%%%%%%

\end{document}

%% file: waste_detection.tex
\title{Waste Detection and Change Analysis based on Multispectral Satellite Imagery}

\titlerunning{Waste Detection and Change Analysis based on Multispectral Satellite Img.}

\author{Dávid Magyar\inst{1}   \and
        Máté Cserép\inst{1}  \and
        Zoltán Vincellér\inst{1} \and
        Attila D. Molnár\inst{2}
}

\authorrunning{Dávid Magyar et al.}

\tocauthor{Dávid Magyar,
           Máté Cserép,
           Zoltán Vincellér,
           Attila D. Molnár
}

\institute{
ELTE Eötvös Loránd University, Faculty of Informatics, Geoinformatics Laboratory \\
\email{\{aoloyk,mcserep,vzoli\}@inf.elte.hu}           % corresponds to \inst{1}
\and
Tisza Plastic Cup Initiative \\
\email{molnar.attila.david@termeszetfilm.hu}           % corresponds to \inst{2}
}

\maketitle

\begin{abstract}
One of the biggest environmental problems of our time is the increase in illegal landfills in forests, rivers, on river banks and other secluded places. In addition, waste in rivers causes damage not only locally, but also downstream, both in the water and washed ashore. Large islands of waste can also form at hydroelectric power stations and dams, and if they continue to flow, they can cause further damage to the natural environment along the river. Recent studies have also proved that rivers are the main source of plastic pollution in marine environments. Monitoring potential sources of danger is therefore highly important for effective waste collection for related organizations. In our research we analyze two possible forms of waste detection: identification of hot-spots (i.e. illegal waste dumps) and identification of water-surface river blockages. We used medium to high-resolution multispectral satellite imagery as our data source, especially focusing on the Tisza river as our study area. We found that using satellite imagery and machine learning are viable to locate and to monitor the change of the previously detected waste.
\end{abstract}

\section{Introduction}
As a result of stricter waste management regulations in the EU, the number of reported waste-related misdemeanors is increasing every day. A significant amount of plastic stocks endanger both human and environmental health and can also lead to water pollution and animal diseases \cite{gabor2019tisza,petpalackaradat,plasticbottletsunami}.
Previous studies in the past decade have identified running freshwaters like rivers as the main source of plastic pollution in marine environments \cite{lechner2014danube}.
It is estimated that approximately 80\% of plastic pollution is carried by rivers into the sea \cite{schmidt2017export}, and the annual plastic input into marine environments from rivers is between 1.15 and 2.41 million metric tons worldwide \cite{lebreton2017river}.
In our research we have focused on Hungary and the surrounding drainage basin territory. While an overall survey on approximately 1000 rivers showed that the average plastic load of the Danube (the major river in Hungary) is well exceeded by rivers in Asia \cite{meijer2021more}, the current state of plastic pollution is still highly problematic. The plastic transport of the Danube river is estimated to 4.2 metric tons / day, so the yearly plastic load to the Black Sea is 1500 tons minimum \cite{lechner2015discharge}. Another long term observations of Hungarian water authorities revealed on the river Tisza the frequency of floating bottles can reach 500 bottles per minute leading to the name \emph{plastic flood} and leaving river Tisza one of the most polluted tributary of the Danube \cite{gyalai2019plastic}.

Therefore it is important that waste collection organizations are able to monitor potential sources of danger. Traditional methods of identifying illegal waste deposits usually require manual surveys, which is not well scalable due to its high demand of human workforce. Remote sensing studies of waste management have so far been underexplored field, due to the spectrally variable and complex nature of different materials (include plastic), and their similarity to other land covers such as water and shadow.

Earth observation satellites equipped with medium- and high-resolution multispectral sensors have undergone great advancement in the past decade. Multitemporal imagery of designated areas became available on a monthly, weekly, or even daily basis, which can also be used for change analysis. These multispectral images can be analyzed based on both their visible and invisible spectrums, as well as the various indices created from them, in order to better separate the waste-covered areas from their surroundings.

Machine learning can be considered a subset of artificial intelligence and consists of computer algorithms which can improve automatically using experience and data. These algorithms build a model based on sample data (so-called training data), which can be utilized to make predictions or decisions without being explicitly programmed to do so. Machine learning algorithms are used in various applications where it is difficult or impractical to develop traditional algorithms to perform the required tasks. Machine learning is widely used nowadays and is increasingly used in remote sensing.

The goal of our research is to develop an accurate classification method for plastic waste detection to provide a viable platform for repeatable, cost-effective and large-scale monitoring. Such a robust waste monitoring solution would speed up the detection of illegal waste hot-spots close to water flows and floating waste islands on rivers, as well as support waste collection actions with an automatic monitoring system. The proposed solution is based on a machine learning approach, the \emph{random forest classification} algorithm \cite{ho1995random}. Some implementations also provide the possibility to control the size of the trees (e.g. the maximum height of the tree or the maximum size of the vertices), which can be used to limit the execution time of the algorithm \cite{rfguidedtour}.

The rest of the paper is structured as follows: Section~\ref{sec:related} introduces the most important concepts related to the topic and provides an insight into waste detection attempts in the literature. Section~\ref{sec:method} presents the areas used to teach the classification model and the detection methods developed during the project. Then Section~\ref{sec:results} illustrates the results obtained. Finally, a summary of the study and our own conclusions are presented in Section~\ref{sec:sum}.

\section{Background and related work}
\label{sec:related}

In recent decades, both remote sensing itself and Earth observation satellites equipped with medium- and high-resolution multispectral sensors have undergone major technological developments. These instruments are capable of recording data in both the visible and invisible range, thus providing more information about the observed area.
In our study we used two Earth observations missions and satellite systems, \emph{Sentinel-2} and \emph{PlanetScope} as data sources.

\subsection{Sentinel-2}
\label{sec:sentinel}

The European Union's Earth observation programme is the Copernicus\footnote{\url{https://www.copernicus.eu/en}} programme \cite{aschbacher2017esa}. Its Sentinel-2 mission is based on a constellation of two identical satellites in the same orbit\footnote{\url{https://sentinels.copernicus.eu/web/sentinel/missions/sentinel-2}}. Each satellite carries an innovative, 13 spectral band, high-resolution multispectral imaging instrument that continuously monitors changes on the Earth's surface.

Sentinel-2's high-resolution multispectral instrument is based on the well-established legacy of the French SPOT\footnote{\url{https://earth.esa.int/eogateway/missions/spot}} and US Landsat satellites\footnote{\url{https://earth.esa.int/eogateway/missions/landsat}}. The multispectral imager is the most advanced of its kind - in fact, it is the first optical earth observation instrument to include three bands in the "red edge" range, which provide key information on the state of vegetation. It integrates two large visible near-infrared and short-wave infrared focal planes, each equipped with 12 detectors and combining 450 000 pixels. Pixels that may fail during the mission can be replaced with redundant pixels. The two types of detectors use high quality filters to ensure perfect separation of spectral bands. The optical-mechanical stability of the instrument must be extremely high, which meant the use of silicon carbide ceramics for the three mirrors and focal plane, as well as for the structure of the telescope itself. The 13 spectral bands, from visible to near infrared and shortwave infrared (as listed in Table~\ref{tab:sentinel_bands}), with a spatial resolution of 10 to 60 metres, bring the observation of the Earth's surface to an unprecedented level of resolution \cite{spoto2012overview,cazaubiel2017multispectral}.

The bands we used were mostly those with a resolution of 10 and 20 meters, with a pixel covering an area of 100 and 400 m$^2$ respectively. Hence, it was only suitable for identifying large landfills.

\begin{table}[htb]
    \centering
    \begin{tabular}{ | c | c | c | l | }
        \hline
        \multicolumn{1}{|m{0.125\textwidth}}{\textbf{Band Number}} & \multicolumn{1}{|m{0.14\textwidth}}{\textbf{Spatial resolution}} & \multicolumn{1}{|m{0.15\textwidth}}{\textbf{Central wavelength}} & \multicolumn{1}{|c|}{\textbf{Description}} \\
        \hline
        \hline
        B1 & 60 m & 443 nm & Coastal aerosol \\
        \hline
        B2 & 10 m & 490 nm & Blue \\
        \hline
        B3 & 10 m & 560 nm & Green \\
        \hline
        B4 & 10 m & 665 nm & Red \\
        \hline
        B5 & 20 m & 705 nm & Vegetation red edge \\
        \hline
        B6 & 20 m & 740 nm & Vegetation red edge	\\
        \hline
        B7 & 20 m & 783 nm & Vegetation red edge \\
        \hline
        B8 & 10 m & 842 nm & NIR \\
        \hline
        B8a & 20 m & 865 nm & Narrow NIR \\
        \hline
        B9 & 60 m & 940 nm & Water vapour \\
        \hline
        B10 & 60 m & 1375 nm & SWIR - Cirrus \\
        \hline
        B11 & 20 m & 1610 nm & SWIR 1 \\
        \hline
        B12 & 20 m & 2190 nm & SWIR 2 \\
        \hline
    \end{tabular}
    \vspace{10pt}
    \caption{Bands and parameters of Sentinel-2 satellite \cite{sentinel2bands,sentinel2bandsnames}.}
    \label{tab:sentinel_bands}
\end{table}

\subsection{PlanetScope}
\label{sec:planet}

The PlanetScope constellation of satellites from Planet Labs Inc. in the US is made up of a multitude of individual cube satellite clusters\footnote{\url{https://earth.esa.int/eogateway/missions/planetscope}}. The constellation of more than 130 satellites is capable of recording almost the entire land area of the Earth on a daily basis. The imaging instrument used on all of these satellites can currently capture images in four bands: \textit{Blue, Green, Red, Near Infrared} (as shown in Table~\ref{tab:planet_bands}), with the addition of a \textit{Red edge} band in special cases.

These instruments were already capable of identifying small areas, as their bands have a resolution of 3 m, i.e. a pixel contains data for an area of 9 m$^2$.

\begin{table}[htb]
    \centering
    \begin{tabular}{ | c | c | c | l | }
        \hline
        \multicolumn{1}{|m{0.12\textwidth}}{\textbf{Band number}} & \multicolumn{1}{|m{0.14\textwidth}}{\textbf{Spatial resolution}} & \multicolumn{1}{|c|}{\textbf{Wavelength}} &  \multicolumn{1}{c|}{\textbf{Description}} \\
        \hline
        \hline
        B1 & 3 m & 455 - 517 nm & Blue \\
        \hline
        B2 & 3 m & 500 - 590 nm & Green \\
        \hline
        B3 & 3 m & 590 - 682 nm & Red \\
        \hline
        B4 & 3 m & 780 - 888 nm & NIR \\
        \hline
    \end{tabular}
    \vspace{10pt}
    \caption{Bands and parameters of PlaneScope satellite \cite{planetscopespec}.}
    \label{tab:planet_bands}
\end{table}

\subsection{Spectral indices}

Different index values can be calculated from the bands of the satellite images. The indices shown in Table~\ref{tab:indices_1} were computed and used in our evaluation. Each one highlights characteristics of different types of areas. Perhaps the most important index we used was the \textit{Plastic Index}, to be described in detail in subsection~\ref{sec:pi}. This index has a higher value for areas containing plastic, which feature we used in the change analysis.

\renewcommand{\arraystretch}{1.4}
\begin{table}[htb]
    \centering
    \begin{tabular}{ l c c }
        $Plastic\ Index\ (PI)$ & $=$ & $\frac{NIR}{NIR\ +\ Red}$ \\
        $Normalized\ Difference\ Water\ Index\ (NDWI)$ & $=$ & $\frac{Green\ -\ NIR}{Green\ +\ NIR}$ \\
        $Normalized\ Difference\ Vegetation\ Index\ (NDVI)$ & $=$ & $\frac{NIR\ -\ Red}{NIR\ +\ Red}$ \\
        $Reversed\ Normalized\ Difference\ Vegetation\ Index\ (RNDVI)$ & $=$ & $\frac{Red\ -\ NIR}{Red\ +\ NIR}$ \\
        $Simple\ Ratio\ (SR)$ & $=$ & $\frac{NIR}{Red}$ \\
    \end{tabular}
    \vspace{10pt}
    \caption{Calculation methods for the used indices \cite{themistocleous2020investigating}.}
    \label{tab:indices_1}
\end{table}
\renewcommand{\arraystretch}{1.0}

\subsection{Plastic Index}
\label{sec:pi}

\begin{figure}[htp]
	\centering
	\includegraphics[width=0.95\textwidth]{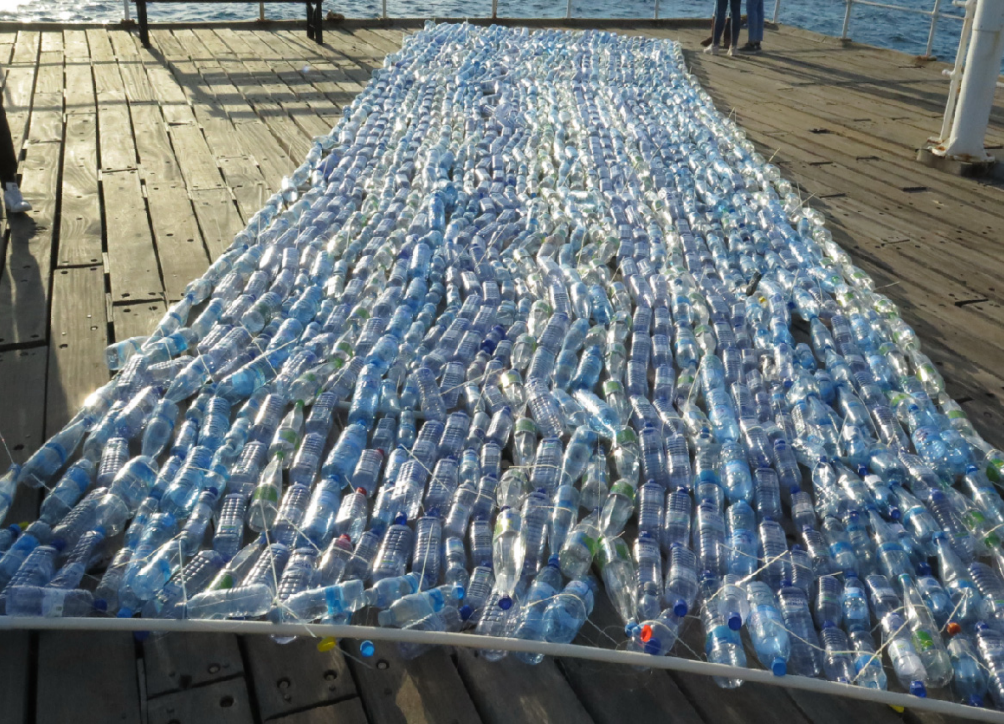}
	\caption{The target made from plastic bottles by Themistocleous et al.}
	\label{fig:plastic_target_1}
\end{figure}

\begin{figure}[p]
	\centering
	\includegraphics[width=0.95\textwidth]{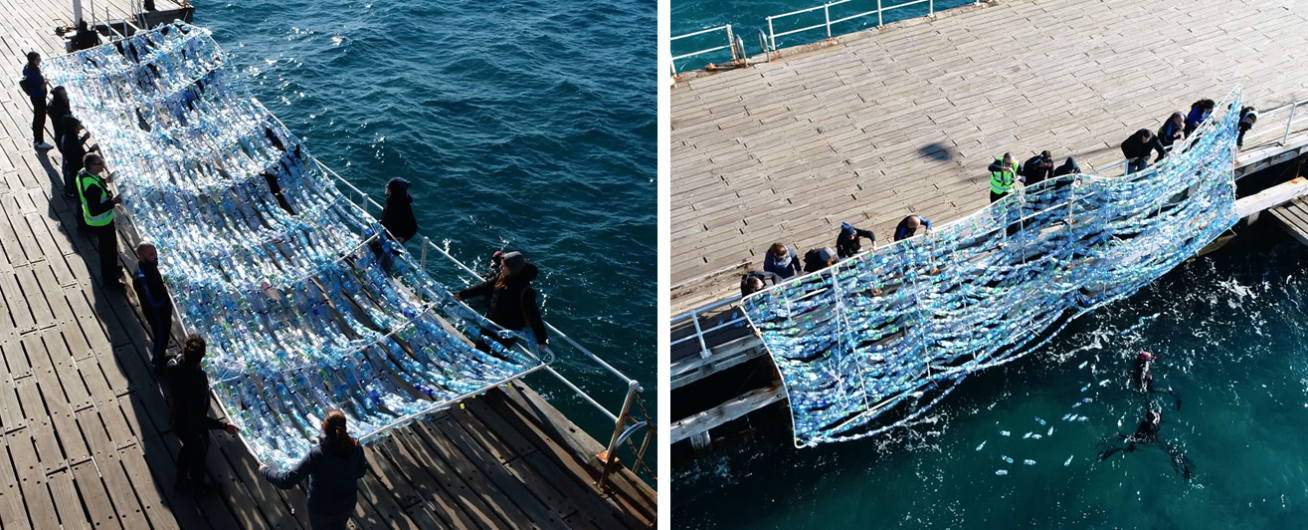}
	\caption{Launching the target by Themistocleous et al.}
	\label{fig:plastic_target_2}
\end{figure}

\begin{figure}[p]
	\centering
	\includegraphics[width=0.95\textwidth]{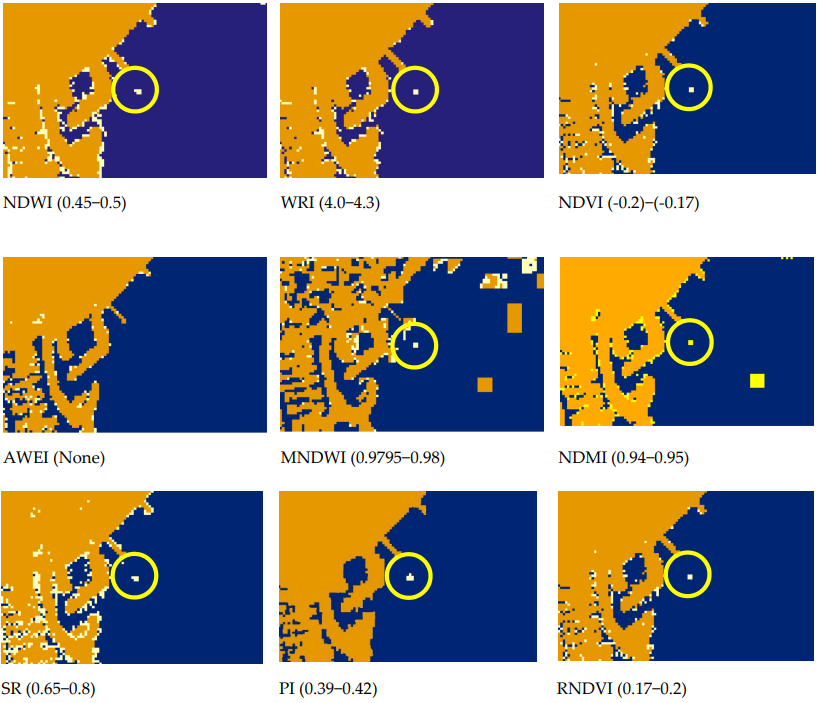}
	\caption{Detection of the target. Sentinel-2 satellite image processed with indices. Land is shown in orange, water in blue and plastic in yellow. The yellow square inside the yellow circle is the plastic target during the satellite pass. The index values for the plastic are shown in brackets for each index \cite{themistocleous2020investigating}.}
	\label{fig:plastic_target_3}
\end{figure}

We have based the theoretical background of waste segmentation on the \emph{Plastic Index}, as introduced by Themistocleous et al. in their work.
In the article the authors investigated whether Sentinel-2 satellite imagery could be used to identify and track plastic litter floating on the sea surface \cite{themistocleous2020investigating}. To do this, a pilot study was carried out. A 3 m × 10 m target consisting of plastic water bottles was created and subsequently placed in the sea near a port in Limassol, Cyprus (see Figures~\ref{fig:plastic_target_1} and~\ref{fig:plastic_target_2}). An unmanned aerial vehicle (UAV) was also used to take multispectral aerial photographs of the study area in the same time as a Sentinel-2 satellite. The spectral signature of the water and the plastic debris placed in the water was recorded using an SVC HR1024 spectroradiometer \cite{svchr1024}. The study found that plastic debris was most easily detected in the Near Infrared (NIR) wavelength. Seven of the indices commonly used in processing satellite imagery were tested for their ability to identify plastic litter in water. In addition, the authors tested two new indices: the \emph{Plastic Index (PI)} and the \emph{Reversed Normalized Difference Vegetation Index (RNDVI)}. The results are visualized in Figure~\ref{fig:plastic_target_3}. The newly developed Plastic Index was able to identify plastic objects floating on the water surface and was the most effective index in identifying plastic litter. In reality, the results may be spoiled by the fact that the bottles are not in such a nice condition, which could change the spectral values.

\subsection{Other attempts for waste detection}
Significant amounts of tyre and plastic waste are a threat to both human and environmental health, can lead to water pollution and animal disease, and are potential sources of fire hazards. Remote sensing studies of waste management have so far been less explored due to the spectrally variable and complex nature of tyres and plastics and their similarity to other land cover such as water and shadows. Therefore, the overall objective of Page et al. was to develop an accurate classification method for detecting both tire and plastic debris to provide a viable platform for repeatable, cost-effective, and large-scale monitoring \cite{page2020identification}. An extended land cover classification combining the thematic indices of the Copernicus Sentinel-2 optical imagery and the Copernicus Sentinel-1\footnote{\url{https://sentinels.copernicus.eu/web/sentinel/missions/sentinel-1}} microwave data has been developed. They used two Random Forest classification algorithms specifically trained for the detection of rubber and plastic debris based on imagery from Scotland.

An important success and application of deep learning in recent years has been in the field of image processing. In their study, Youme et al. present an automated solution for detecting hidden landfills utilizing unmanned aerial vehicle (UAV) imagery in the Saint Louis area of Senegal, West Africa\footnote{Saint Louis, Senegal, West Africa -- GPS coordinates (lat/long): 16.0234 / -16.4792}, using a convolutional neural network (CNN) \cite{youme2021deep}. The results show that the model is good at identifying the areas concerned, but has difficulties in some areas where there are no clear ground truths.

\section{Methodology}
\label{sec:method}

As introduced before, two waste detection methods were analyzed in our study:
\begin{enumerate}
    \item Identification of hot-spots, i.e. illegal waste dumps on medium to high-resolution satellite images in the upper parts of the river Tisza (Ukraine, Romania), from which significant amounts of waste are discharged into the river during floods.
    \item Identification of water-surface river blockages on medium to high-resolution satellite imagery that can form at known locations on the river (e.g. hydroelectric dams).
\end{enumerate}

\subsection{Identification of hot-spots}
The aim was to be able to observe the changes in the extent of the polluted areas along the river. Such areas directly on the river banks are a major hazard during floods. In such cases, the river leaves its bed and when it recedes, it carries the waste with it and deposits it in another, undesired location. Such hot-spots along the river Tisza are mainly located in Ukraine and Romania. Hungarian waste collection organizations already have experience of the hot-spots where they have to move to after a flood. If our classification shows that the extent of the hot-spot is significant in the pre-flood picture, then an on-site investigation is justified.

\begin{figure}[htp]
	\centering
	\subcaptionbox{Before classification}{
		\includegraphics[width=0.45\linewidth]{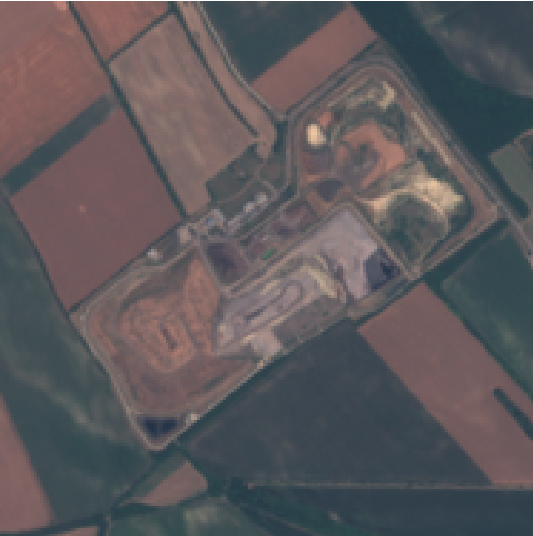}}
	\hspace{5pt}
	\subcaptionbox{After classification}{
		\includegraphics[width=0.45\linewidth]{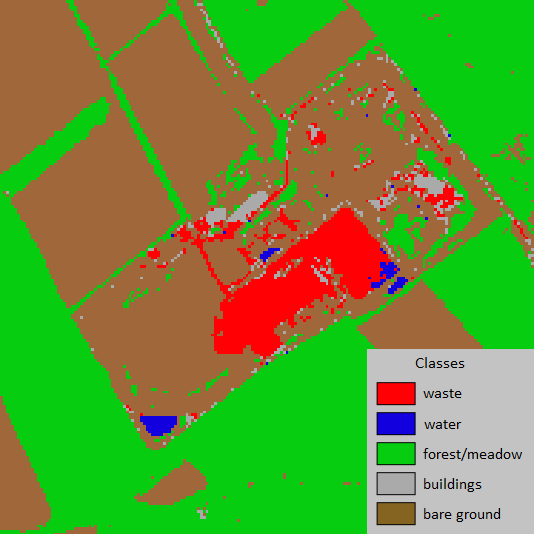}}
	\caption{Classification of the teaching area in Pusztazámor.}
	\label{fig:pusztazamor_classification}
\end{figure}

The landfill in Pusztazámor played a major role in teaching the model used for this method, as the aim was to detect features very similar to this area. No other procedure than classification was used. The results are visualized in Figure~\ref{fig:pusztazamor_classification}.

\begin{figure}[htp]
	\centering
	\subcaptionbox{Before classification}{
		\includegraphics[width=0.275\linewidth]{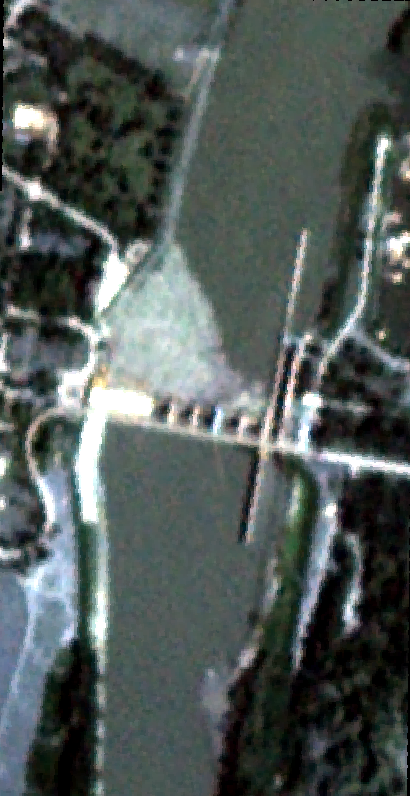}}
	\hspace{5pt}
	\vspace{10pt}
	\subcaptionbox{After classification}{
		\includegraphics[width=0.275\linewidth]{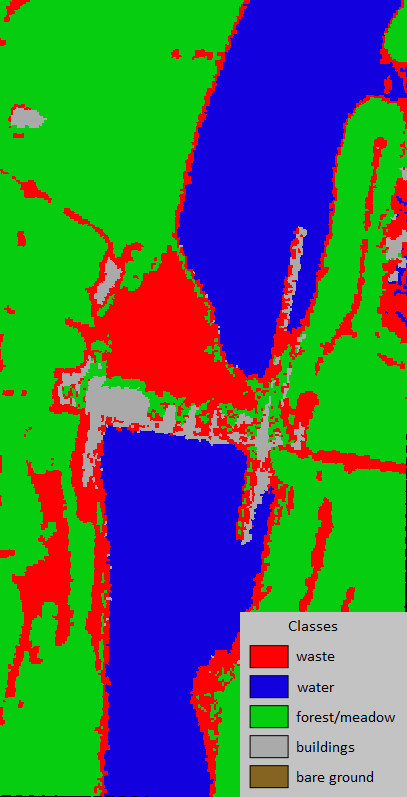}}
	\hspace{5pt}
	\subcaptionbox{Binary image}{
		\includegraphics[width=0.275\linewidth]{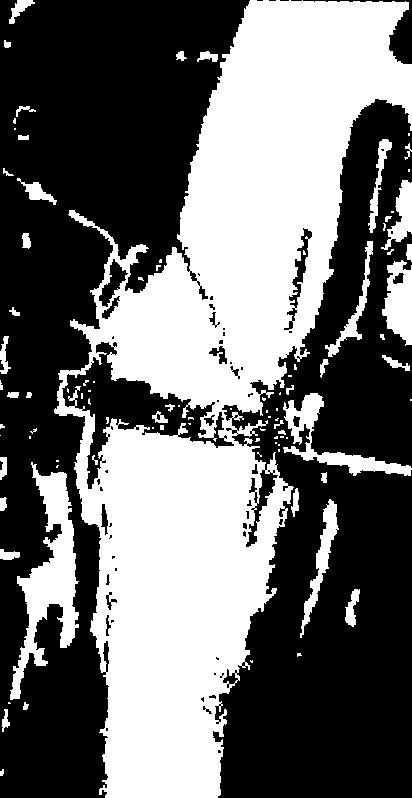}}
	\hspace{5pt}
	\subcaptionbox{Opening}{
		\includegraphics[width=0.275\linewidth]{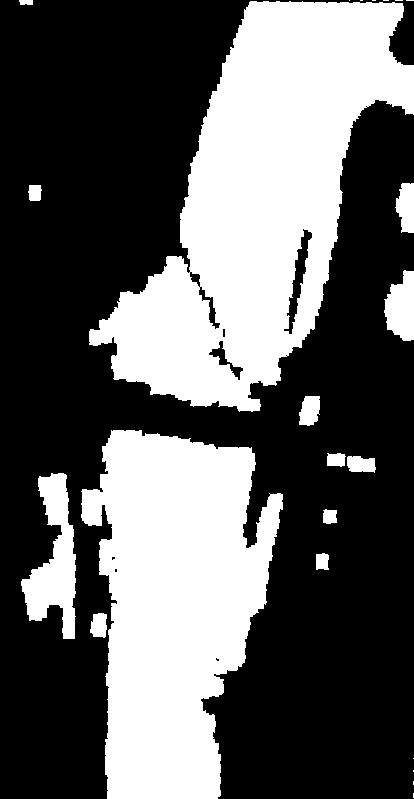}}
	\hspace{5pt}
	\subcaptionbox{Dilation}{
		\includegraphics[width=0.275\linewidth]{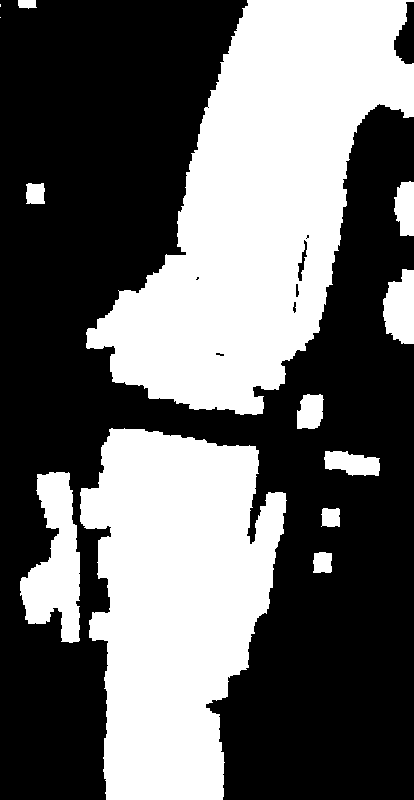}}
	\hspace{5pt}
	\subcaptionbox{Result}{
		\includegraphics[width=0.275\linewidth]{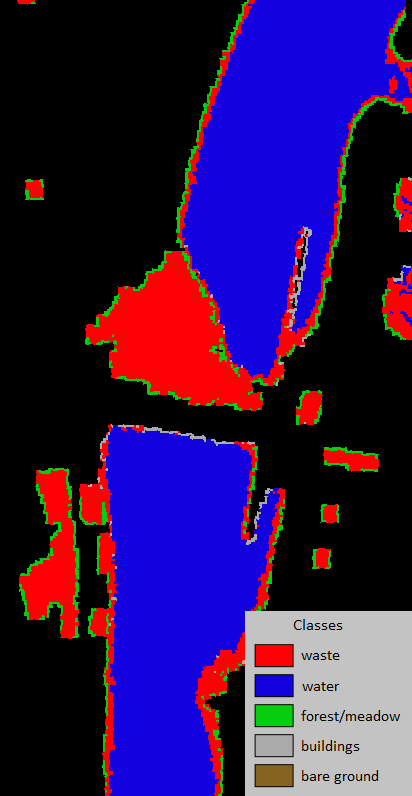}}
	\caption{Detection of the garbage barrage at the hydropower plant in Kisköre.}
	\label{fig:kiskore_process}
\end{figure}

\subsection{Identification of water-surface river blockages}

Water-surface river blockages can form mainly at hydroelectric power plants or dams. Such a barrage can cover up to several thousand square metres. The images of the hydropower plant in Kisköre were a great help in detecting them, because large areas of rubbish often get stuck there.

In addition to classification, morphological transformations were also needed to achieve the desired results. From the classified image, we created a binary one: waste and water classes were grouped together and everything else was left out. We performed a morphological opening on the result in order to remove noise (small areas that fall away from the river). This was followed by a dilation to widen the contours of the remaining areas. For these we used a 5 × 5 matrix as a kernel. The end result seen in Figure~\ref{fig:kiskore_process} is now mainly just the river and the island of junk floating on it.

\subsection{The model}

As mentioned before, the teaching areas were the landfill in Pusztazámor\footnote{Landfill (Pusztazámor) -- GPS coordinates (lat/long): 47.3778 / 18.7982} and the hydroelectric power plant in Kisköre\footnote{Hydroelectric power plant (Kisköre) -- GPS coordinates (lat/long): 47.4935 / 20.5150}. These sites are covered with large quantities of waste visible to the naked eye on satellite images. We used Sentinel-2 images of the landfill and PlanetScope images of the power plant.

Two different methods were used in the teachings. In one, we used only the Sentinel-2 images and indices that could be calculated from the available bands (Table~\ref{tab:sentinel_bands}). In the other version, we used only those bands that are possible to obtain from both Sentinel-2 and PlanetScope images (Table~\ref{tab:planet_bands}). The reason for this was that we wanted to embed the final solution in our own framework, which is able to process Sentinel-2 and PlanetScope images in a consistent way. From these bands we calculated the indices presented in Table~\ref{tab:indices_1}.

The final model uses five target classes: waste, water, forest/meadow, buildings and bare ground. We also performed cross-validation on almost 200,000 training data, which showed an accuracy of almost 96\%.

\section{Results}
\label{sec:results}

This chapter presents some of the results of the developed methods. Where possible, several sample areas are illustrated. The classification of the autumn and winter recordings did not give good results. The reason being that due to bad weather conditions no recordings were made that could be meaningfully used for teaching. The results shown below were obtained by processing spring and summer images.

In the classified images, the most important color is red, as this indicates the areas covered by waste according to the model. The heat maps show a subset of the littered areas. The colors on these represent how confident the model was in its decision (see Table~\ref{tab:heatmap_colors}).

\definecolor{red}{RGB}{255,0,4}
\definecolor{blue}{RGB}{18,0,222}
\definecolor{green}{RGB}{6,205,16}
\definecolor{gray}{RGB}{170,170,170}
\definecolor{brown}{RGB}{133,100,33}
\definecolor{yellow}{RGB}{246,221,0}

\begin{table}[htb]
    \begin{minipage}{.5\linewidth}
        \centering
        \begin{tabular}{ | c | c | m{0.05\textwidth} | }
            \hline
            \multicolumn{1}{|c|}{\textbf{Class}} &
            \multicolumn{2}{|c|}{\textbf{Color}} \\
            \hline
            \hline
            waste & red & \cellcolor{red} \\
            \hline
            water & blue & \cellcolor{blue} \\
            \hline
            forest/meadow & green & \cellcolor{green} \\
            \hline
            buildings & gray & \cellcolor{gray} \\
            \hline
            bare ground & brown & \cellcolor{brown} \\
            \hline
        \end{tabular}
        \vspace{10pt}
        \caption{Classes and their coloring.}
        \label{tab:classification_colors}
    \end{minipage}%
    \begin{minipage}{.5\linewidth}
       \centering
        \begin{tabular}{ | c | c | m{0.05\textwidth} | }
            \hline
            \multicolumn{1}{|c|}{\textbf{Confidence}} &
            \multicolumn{2}{|c|}{\textbf{Color}} \\
            \hline
            \hline
            90\% - 100\% & red & \cellcolor{red} \\
            \hline
            80\% - 90\% & yellow & \cellcolor{yellow} \\
            \hline
            70\% - 80\% & green & \cellcolor{green} \\
            \hline
        \end{tabular}
        \vspace{10pt}
        \caption{Coloring of heat maps.}
        \label{tab:heatmap_colors}
    \end{minipage} 
\end{table}

\subsection{Identification of hot-spots}

\begin{figure}[htp]
	\centering
	\subcaptionbox{Before process}{
		\includegraphics[width=0.7\linewidth]{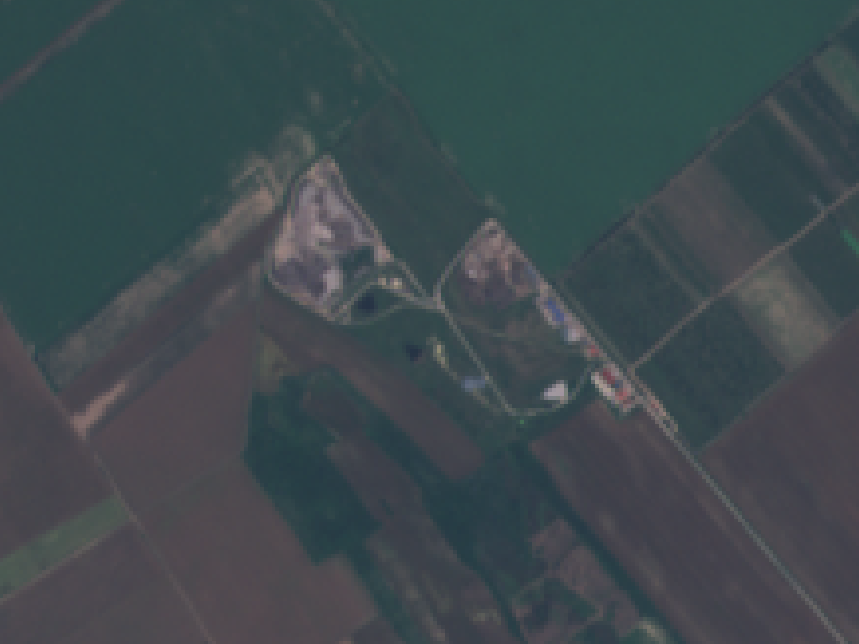}}
	\subcaptionbox{After process}{
		\includegraphics[width=0.7\linewidth]{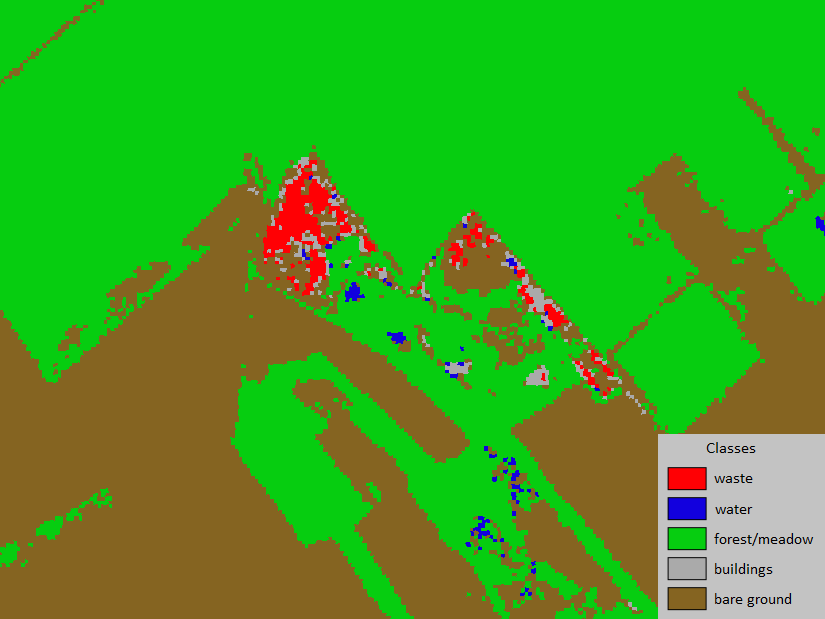}}
	\subcaptionbox{Heatmap}{
		\includegraphics[width=0.7\linewidth]{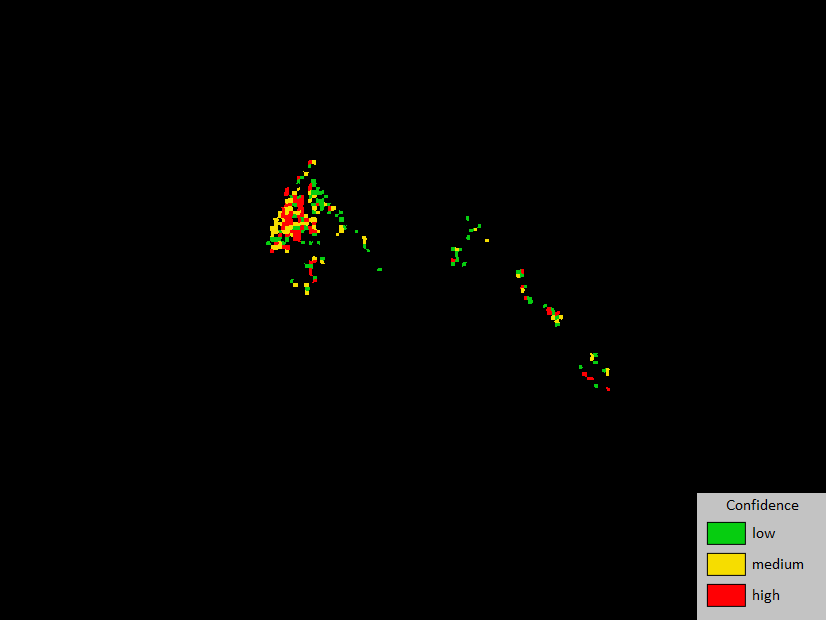}}
	\caption{Deponia Waste Management Centre, Hungary - June 27, 2019}
	\label{fig:hotspot_result_1}
\end{figure}

\begin{figure}[htp]
	\centering
	\subcaptionbox{Before process}{
		\includegraphics[width=0.7\linewidth]{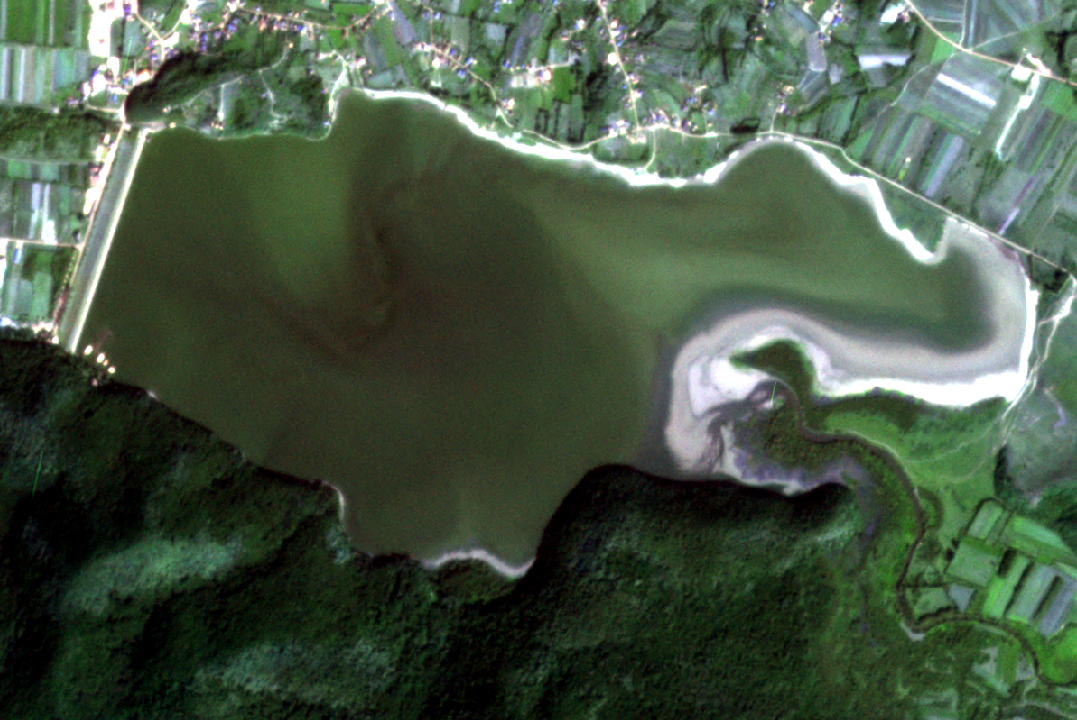}}
	\hspace{5pt}
	\subcaptionbox{After process}{
		\includegraphics[width=0.7\linewidth]{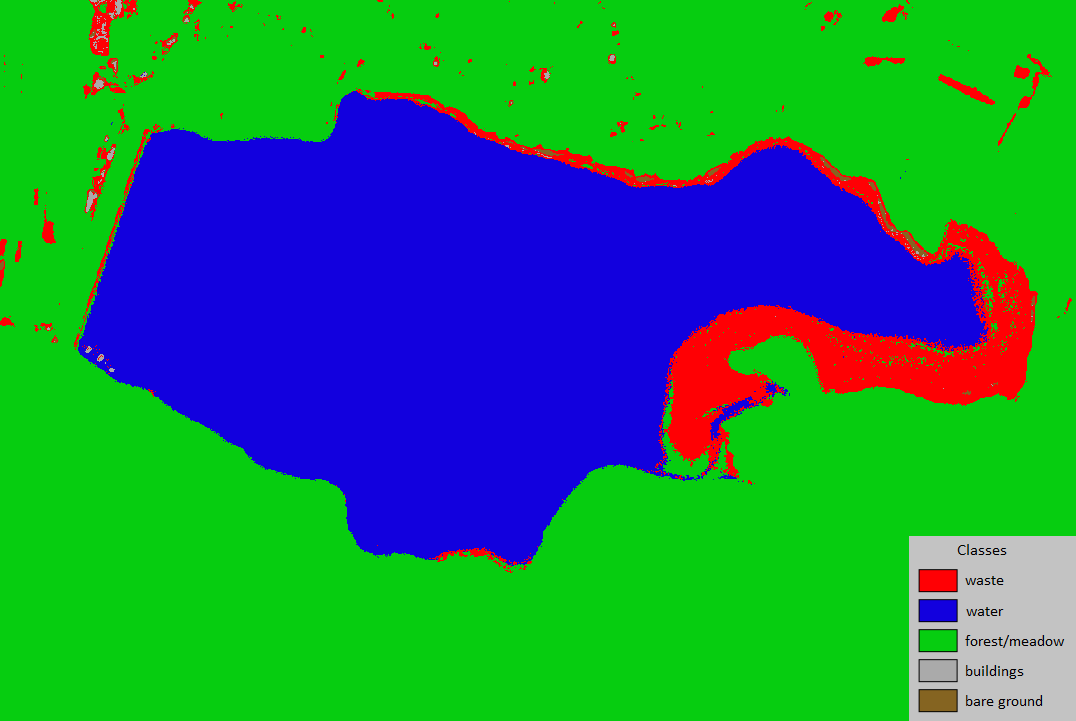}}
	\hspace{5pt}
	\subcaptionbox{Heatmap}{
		\includegraphics[width=0.7\linewidth]{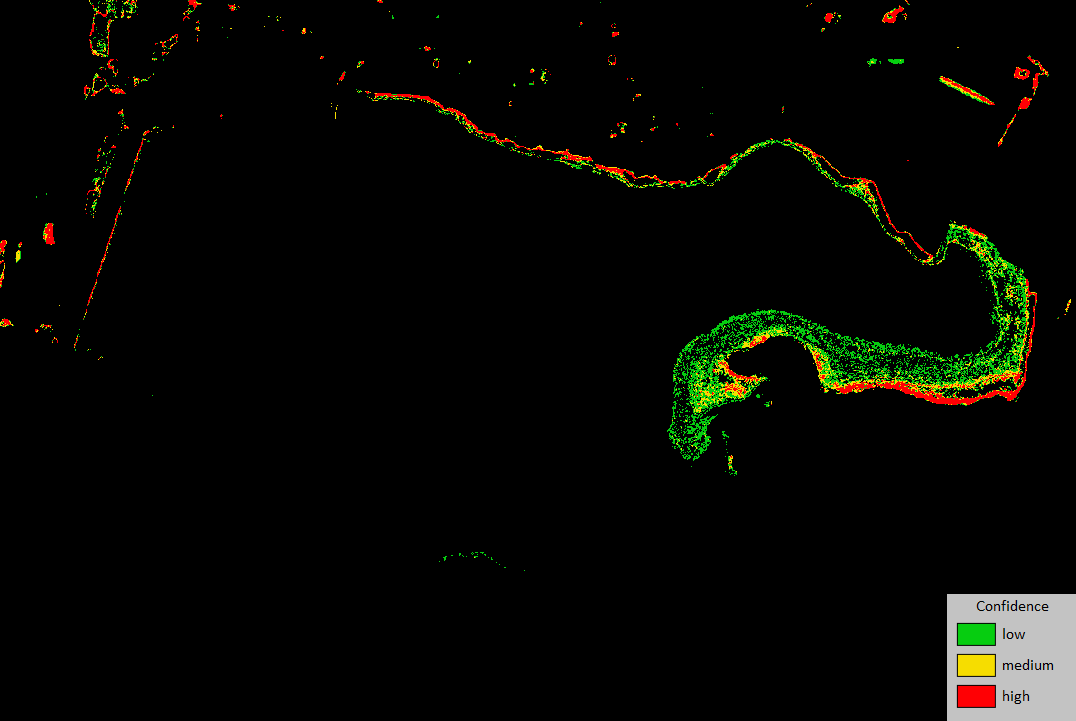}}
	\caption{Lake Călinești, Romania - September 16, 2020}
	\label{fig:hotspot_result_2}
\end{figure}

In this method, the process consisted of only one classification. Figure~\ref{fig:hotspot_result_1} shows the classification of the area of the Deponia Waste Management Centre\footnote{Deponia Waste Management Centre -- GPS coordinates (lat/long): 47.2399 / 18.4682} in Hungary, and Figure~\ref{fig:hotspot_result_2} shows one of the areas identified by the Plastic Cup staff, Lake Călinești\footnote{Lake Călinești -- GPS coordinates (lat/long): 47.8946 / 23.2948} in Romania, which is considered a potential hot-spot.

\subsection{Identification of water-surface river blockages}

\begin{figure}[htp]
	\centering
	\subcaptionbox{Before process}{
		\includegraphics[width=0.3\linewidth]{resources/Kiskore_20190702_RGB.png}}
	\hspace{5pt}
	\subcaptionbox{After process}{
		\includegraphics[width=0.3\linewidth]{resources/Kiskore_20190702_MASKED_CLASSIFIED-overlay.png}}
	\hspace{5pt}
	\subcaptionbox{Heatmap}{
		\includegraphics[width=0.3\linewidth]{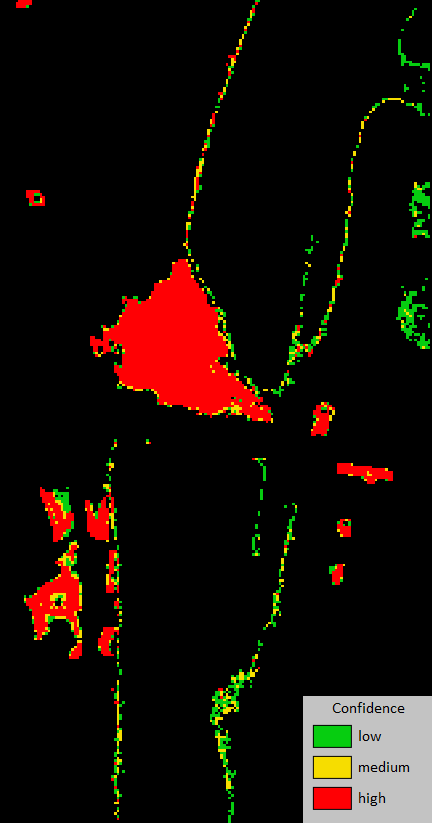}}
	\caption{Hydropower Plant in Kisköre, Hungary - July 2, 2019}
	\label{fig:floating_result_1}
\end{figure}

\begin{figure}[htp]
	\centering
	\subcaptionbox{Before process}{
		\includegraphics[width=0.3\linewidth]{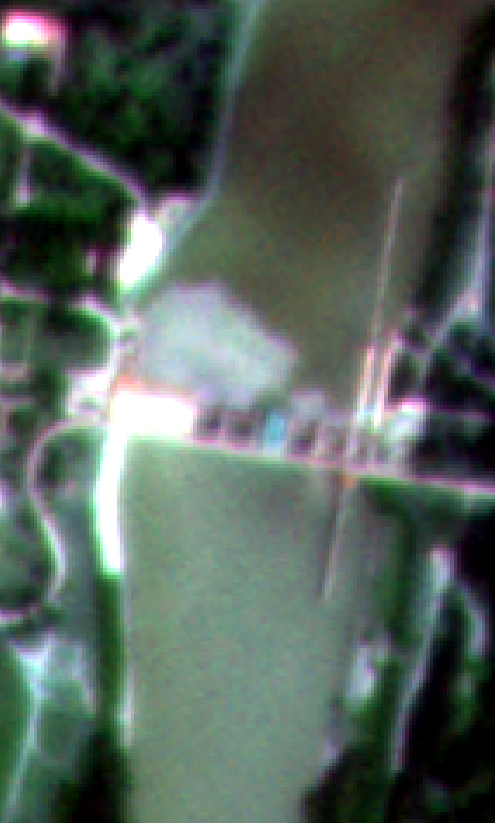}}
	\hspace{5pt}
	\subcaptionbox{After process}{
		\includegraphics[width=0.3\linewidth]{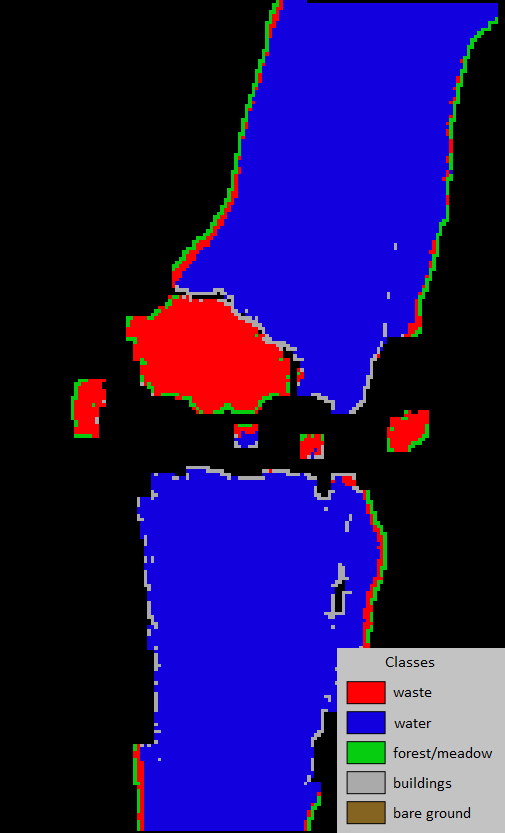}}
	\hspace{5pt}
	\subcaptionbox{Heatmap}{
		\includegraphics[width=0.3\linewidth]{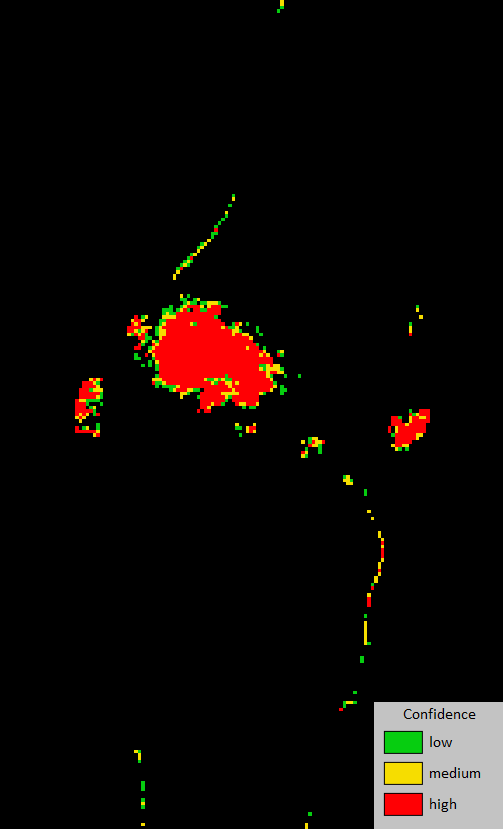}}
	\caption{Hydropower Plant in Kisköre, Hungary - July 23, 2020}
	\label{fig:floating_result_2}
\end{figure}

This method gave the most promising results. The areas of detected waste (shown in red) in the classified images are clearly separated from their surroundings, as presented in Figures~\ref{fig:floating_result_1} and~\ref{fig:floating_result_2}. We considered eliminating the red spots on the river banks that are not in contact with water (blue), but concluded to leave them in the final images. The reason for this was that there have been cases where the garbage island was in contact with other classes rather than water on the classified image. In this case the actual waste would have been removed.

\subsection{Visualizing the results}

\begin{figure}[htp]
	\centering
	\includegraphics[width=\textwidth]{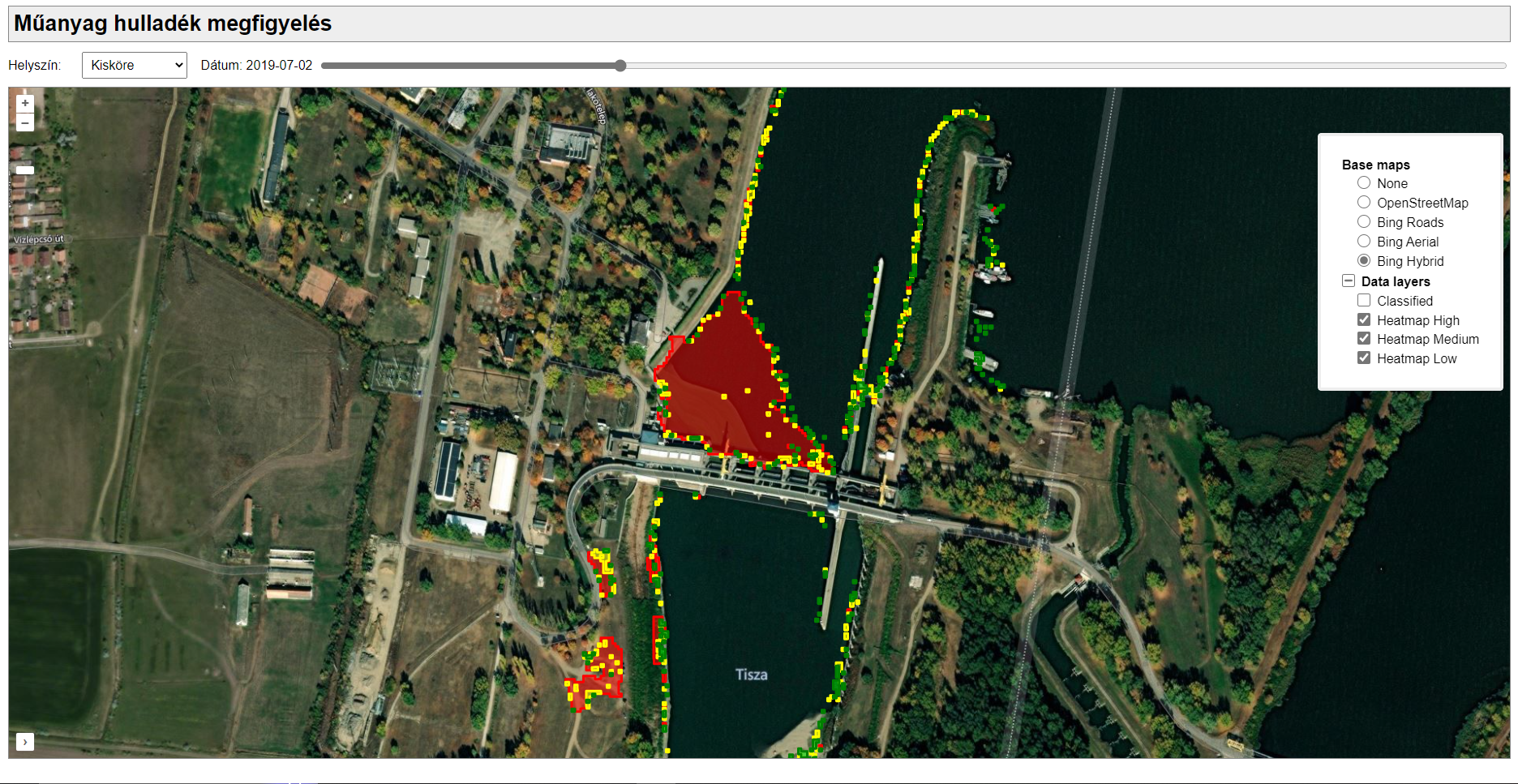}
	\caption{User interface of web application for visualizing the results.}
	\label{fig:web_app}
\end{figure}

An automated evaluator for waste monitoring and change detection was also developed as part of the research. The program provides an easy and robust solution to configure territories for continuous observation. The application downloads satellite images from the mentioned sources on a daily basis (given that a new image is available for the selected area) and compares the amount of waste covered surface to previous images. In case there is a noticeable change, an alert is sent to preconfigured email addresses.
We also started to develop a web application that provides information about the status of predefined locations. The goal is to visualize the extent of polluted areas in these locations. In the final version it will be updated on a daily basis, showing only the latest results. In Figure~\ref{fig:web_app} you can see the prototype user interface of this application.

% \begin{figure}[ht]
%  \hbox{%
%   \hspace{.025\hsize}%
%   \begin{minipage}[c]{.3\hsize}
%   \epsfig{file=resources/Kiskore_20200723_RGB.png, width=\linewidth}
%   \centerline{(a)}
%   \end{minipage}%
%   \hspace{.025\hsize}%
%   \begin{minipage}[c]{.3\hsize}
%   \epsfig{file=resources/Kiskore_20200723_CLASSIFIED-overlay.png, width=\linewidth}
%   \centerline{(b)}
%   \end{minipage}%
%   \hspace{.025\hsize}%
%   \begin{minipage}[c]{.3\hsize}
%   \epsfig{file=resources/Kiskore_20200723_CLASSIFIED_HEATMAP-overlay.png, width=\linewidth}
%   \centerline{(b)}
%   \end{minipage}%
%  }%\hbox
%  \caption{Kisköre Hydropower Plant, Hungary - 2020.07.23.}
% \label{fig:floating_result_2} 
% \end{figure}

\subsection{Execution time}

Table~\ref{tab:runtime} shows the execution time of our waste detection methods. It should be noted that the significant part of these processes are the calculation of indices. Compared to this, the classification and morphological transformations are computed much faster.

\begin{table}[H]
    \centering
    \begin{tabular}{ | c | c | r | }
        \hline
        \multicolumn{1}{|c|}{\textbf{Process}} &
        \multicolumn{1}{|c|}{\textbf{Image size}} & \multicolumn{1}{|c|}{\textbf{Execution time}} \\
        \hline
        \hline
        \multirow{4}{*}{\textbf{Hot-spots}} & 164 $\times$ 312 = 51 168 &  2.17 sec \\
        \cline{2-3}
        & 1194 $\times$ 801 = 956 394 & 39.98 sec \\
        \cline{2-3}
        & 4597 $\times$ 4153 = 19 091 341 & 12 min 42 sec\\
        \cline{2-3}
        & 6614 $\times$ 5981 = 39 558 334 & 26 min 5 sec\\
        \hline
        \multirow{4}{*}{\shortstack[c]{\textbf{Water-surface}\\ \textbf{river blockages}}} & 164 $\times$ 312 = 51 168 & 2.5 sec \\
        \cline{2-3}
        & 1194 $\times$ 801 = 956 394 & 42.45 sec \\
        \cline{2-3}
        & 4597 $\times$ 4153 = 19 091 341 & 14 min 7 sec \\
        \cline{2-3}
        & 6614 $\times$ 5981 = 39 558 334 & 29 min 41 sec\\
        \hline
    \end{tabular}
    \vspace{5pt}
    \caption{Execution time of the waste detection methods.}
    \label{tab:runtime}
\end{table}

\section{Conclusion}
\label{sec:sum}

The aim of the research was to test and develop waste detection methods that can be used in practice and facilitate the work of waste collection organizations. The solution developed for this purpose is compatible with any satellite image with at least four bands, which must necessarily include the \textit{Blue, Green, Red, Near Infrared} bands.

The trained \textit{Random Forest} classification model works well mainly with images taken in spring and summer, because the weather conditions are not suitable in winter and autumn, when images are usually too cloudy and could not be used in teaching the model.

Future work could include teaching the model using very large amounts of data. Also, the processing of the images could be done in parallel, which could lead to a reduction in processing time. We will continue developing the web application to further facilitate the practical use of this project.

\section*{Acknowledgements}
The research work was supported financially by the ELTE Faculty of Informatics and InforNess Training Ltd. The PlanetScope satellite images were provided by Planet Labs Inc. Education and Research Program \cite{team2017planet}.
Professional support was provided by the Lechner Knowledge Centre\footnote{\url{https://lechnerkozpont.hu/}} and the Tisza Plastic Cup\footnote{\url{https://petkupa.hu/}} waste collection organization.

\bibliographystyle{IEEEtran}
\bibliography{references}